\documentclass[a4paper]{article}

\usepackage[english]{babel}
\usepackage[utf8x]{inputenc}
\usepackage[T1]{fontenc}

\usepackage[a4paper,top=3cm,bottom=2cm,left=3cm,right=3cm,marginparwidth=1.75cm]{geometry}

\usepackage{amsmath}
\usepackage{graphicx}
\usepackage[colorinlistoftodos]{todonotes}
\usepackage[colorlinks=true, allcolors=blue]{hyperref}
\usepackage{authblk}
\usepackage{makecell}

\title{End-to-end 3D shape inverse rendering of different classes of objects from a single input image}

\author[1]{Shima Kamyab}
\author[1]{S. Zohreh Azimifar}
\affil[1]{Computer Science and Engineering and Information Technology, Shiraz university, Shiraz, Iran}
\begin{document}
\maketitle

\begin{abstract}
In this paper a semi-supervised deep framework is proposed for the problem of 3D shape inverse rendering from a single 2D input image. The main structure of proposed framework consists of unsupervised pre-trained components which significantly reduce the need to labeled data for training the whole framework. using labeled data has the advantage of achieving to accurate results without the need to predefined assumptions about image formation process. Three main components are used in the proposed network: an encoder which maps 2D input image to a representation space, a 3D decoder which decodes a representation to a 3D structure and a mapping component in order to map 2D to 3D representation. The only part that needs label for training is the mapping part with not too many parameters. The other components in the network can be pre-trained unsupervised using only 2D images or 3D data in each case. The way of reconstructing 3D shapes in the decoder component, inspired by the model based methods for 3D reconstruction, maps a low dimensional representation to 3D shape space with the advantage of extracting the basis vectors of shape space from training data itself and is not restricted to a small set of examples as used in predefined models. Therefore, the proposed framework deals directly with coordinate values of the point cloud representation which leads to achieve dense 3D shapes in the output. The experimental results on several benchmark datasets of objects and human faces and comparing with recent similar methods shows the power of proposed network in recovering more details from single 2D images.
\end{abstract}

\section{Introduction}

Inverse rendering stands for a specific class of machine learning techniques aim at recovering the properties of a 3D scene like camera’s extrinsic parameters, scene lighting and the shape of the scene from some existing measurements (e.g. 2D images)\cite{aldrian2013inverse, patow2003survey}. It has attracted high interest in the recent academic research due to wide applications in computer vision \cite{kim2016multi,aldrian2013inverse,wang2017ultrasound,wood20163d,zhu2016face,garrido2016reconstruction,richardson20163d,thies2016face2face,tran2016regressing,rezende2016unsupervised,richardson2016learning,tewari2017mofa,jiang20173d,patow2003survey,fouhey2016understanding}.

Inverse rendering is not a well posed problem because of existing infinite number of 3D structure which can result the same 2D image \cite{garrido2016reconstruction, rezende2016unsupervised}. Facing with such problems needs additional assumptions and prior knowledge about the class of 3D structures to be recovered from 2D images. For instance a statistical model can be used to determine which solution is likely and which is not for a specific class of objects \cite{rezende2016unsupervised}.\par
The aim of Inverse rendering problem is actually inverting the image formation process called rendering. in Computer graphics, Rendering is expressed as an equation formulated as in \ref{eq1} \cite{marschner1998inverse}.
\begin{equation} \label{eq1}
F=K(h+Gf)
\end{equation}

where, F is the reflected radiance (light) from the surface (e.g. intensity of the pixels in the image), K represents the way the object surface reflects the lights in various spatial and angular positions, h describes the lighting of the scene, G is some measurement showing how light is transferred between other existing surfaces in the scene and f is the light which is reflected by other surfaces in scene affecting F.
Knowing the quantities K, h, G, f means that the properties of the scene are known and F can be computed using these quantities and this is a direct problem to solve. The inverse rendering problem arises when F or some information about it are available and the aim is to compute some values in the right-hand side of \ref{eq1}. In this paper, our focus is on recovering shape geometry of an object from a single 2D input image having different poses which is implicitly embedded in variable K as spatial properties of the surface in light reflection. \par
Inverse rendering problem as a machine learning task, can be faced through 3 main steps to be solved:
\begin{itemize}
  \item Selection of a model (i.e. assuming a specific class of hypotheses) 
  \item Defining a score criterion
  \item Defining a Search strategy
\end{itemize}

\textbf{Selection of a model:} Selection of a model is one of the main necessities for solving machine learning problems due to “no free lunch theorem” \cite{wolpert1997no}. In the case of recovering 3D structures from 2D images, this step is more important because in such problems we deal with an ill-posed problem based on Hadmard definition for well-posed problems: there might be more than one solutions that give the same 2D image and the solution does not depends continuously on the data, which means that small errors in measurements may cause large errors in the solutions \cite{poggio1985ill}. For solving such kinds of problems, some assumptions are usually made about the properties of the solutions to restrict the solution space to feasible regions. Selection of a suitable model for defining the target scene serves as some prior knowledge about the solution space, by which the search can be performed more reliable. Like using any other assumptions, this method may have some costs like losing some promising regions that are not spanned by the selected model. For inverse rendering problems, there are various types of models and prior knowledge for the target scene in the literature that make the existing methods different from each other: some approaches use additional information, like multiple images or video sequence or landmarks, about the scene to find a unique solution \cite{zhu2016face,piotraschke2016automated,gao2016exploiting,lun20173d}, some of them work with a certain statistical model to determine the feasibility of the solutions \cite{jackson2017large} and some other perform some regressing techniques that relates the output measurements to 3D structures \cite{garrido2016reconstruction,richardson20163d,thies2016face2face,tran2016regressing,rezende2016unsupervised,richardson2016learning,jiang20173d,liu2017interactive,tewari2017mofa}.\par
In this paper our focus is on the pose invariant recovering 3D structure of an object from a single 2D image with the help of a model describing the properties of that object. The main structure of our proposed framework is based on finding a relation between measurements, i.e. a single 2D image for each scene, and 3D structure of the scene constrained with the used model. \par
\textbf{Defining a score criterion:}After selection of a suitable model describing the solution space, the optimum solution should be searched in that space. For this aim, first a criterion should be defined which guides the search toward finding the optimum solution satisfying the constraints of the problem at hand. For instance, in the case of inverse rendering of human face from 2D image(s) for face recognition task, first, the main objective is to recover the distinctiveness characteristics of a face in an image and so a good solution should not be too generic so or too over determined and there should be some tradeoff between these two properties in the objective criterion \cite{tran2016regressing}. In this paper, since the objective is to minimize the reconstruction error, we used Root Mean Square Error (RMSE) as loss function for training and evaluation. \par

\textbf{Defining a search strategy:}There exist different types of search mechanisms to search for a solution based on the defined score criterion. In the case of convex criteria, a closed form solution exists and the search can be done in linear time. However, the nature of inverse rendering problem is highly nonlinear and the inference and finding the relation between 2D image and corresponding 3D structure is intractable. If some assumptions can be incorporated to the problem, like deterministic models formulating behavior of all variables in the rendering equation to be defined accurately, so the solution can be computed in closed form \cite{aldrian2013inverse}. However we cannot have such strong assumptions in all inverse rendering problems. In such cases we can use learning based methods by training some models and then use them for test samples. The most popular learning framework in recent years are deep networks which show their power in solving nonlinear problems. Deep networks like other learning based structures can be trained in 3 ways: Supervised, unsupervised and semi-supervised. \par
Supervised methods use a set of labeled data to guide the training. Their result will be accurate because of availability of ground truth in the training procedure but suffer from the lack of enough labeled training data in some domains. In the case of 3D Inverse rendering applications, providing enough realistic training data for a deep network may be impractical. One Solution for this problem is to use a generative model for generating synthetic training data to trained the network \cite{richardson20163d,richardson2016learning}. \par
As another solution, There can be found some attempts in the recent literature to Unsupervised training the network for 3D reconstruction\cite{rezende2016unsupervised,tewari2017mofa} in which a 3D structure is generated in a part of the network and some rendering mechanism is used to transform the generated 3D structure to something in the form of input measurement to the network and then try to reduce the difference between input and rendered output. So the network does not need to labeled data for training. In the existing unsupervised methods in the literature, they used an analytical form for rendering the 3D structure with some assumptions about image formation process. This characteristic although is a significant improvement in reducing the need for training data but may affects the quality of solutions due to the existing assumptions about rendering mechanism (such as camera properties, scene illumination, reflectance properties and so on). In the case of 3D Inverse rendering problem, there are well known 3D models that can help us to generate enough data for training the deep networks.\par
In this paper, we propose a mechanism in which some parts of network can be trained using unlabeled data and then the whole network will be fine-tuned using labeled training data for aggregation of the components. We believe this properties reduces the nedd of deep network for large set of training data. Therefore it is a semi-supervised framework for inverse rendering problem.\par
The paper is organized as follows: we first briefly describe our framework in section \ref{s1}. In section \ref{s2} we review and analyze some recent related works. The proposed framework and network structure will be define in more details in section \ref{s3}. Some experimental results to verify the performance of our proposed structure on several benchmarks will be demonstrated in section \ref{s4} and finally some discussions and conclusions will be found in sections \ref{s5} and \ref{s6}.

\section{Overview}\label{s1}
Our proposed framework is composed of 3 main components based on the fact that in inverse rendering problem we look for a way of transforming the 2D input image to its corresponding 3D structure. We do this by mapping the 2D input image and corresponding 3D structure to suitable representations and then finding a transformation between obtained representations. We will show experimentally that nonlinear mapping of suitable representations results more accurate solutions. finding suitable representations can be done using autoencoders as unsupervised tools for finding representations or by pre-trained convolutional networks for finding suitable representations. We did this either by different types of autoencoder or by some existing well known networks from literature.\par
\begin{figure}
\centering
\includegraphics[width=1\textwidth]{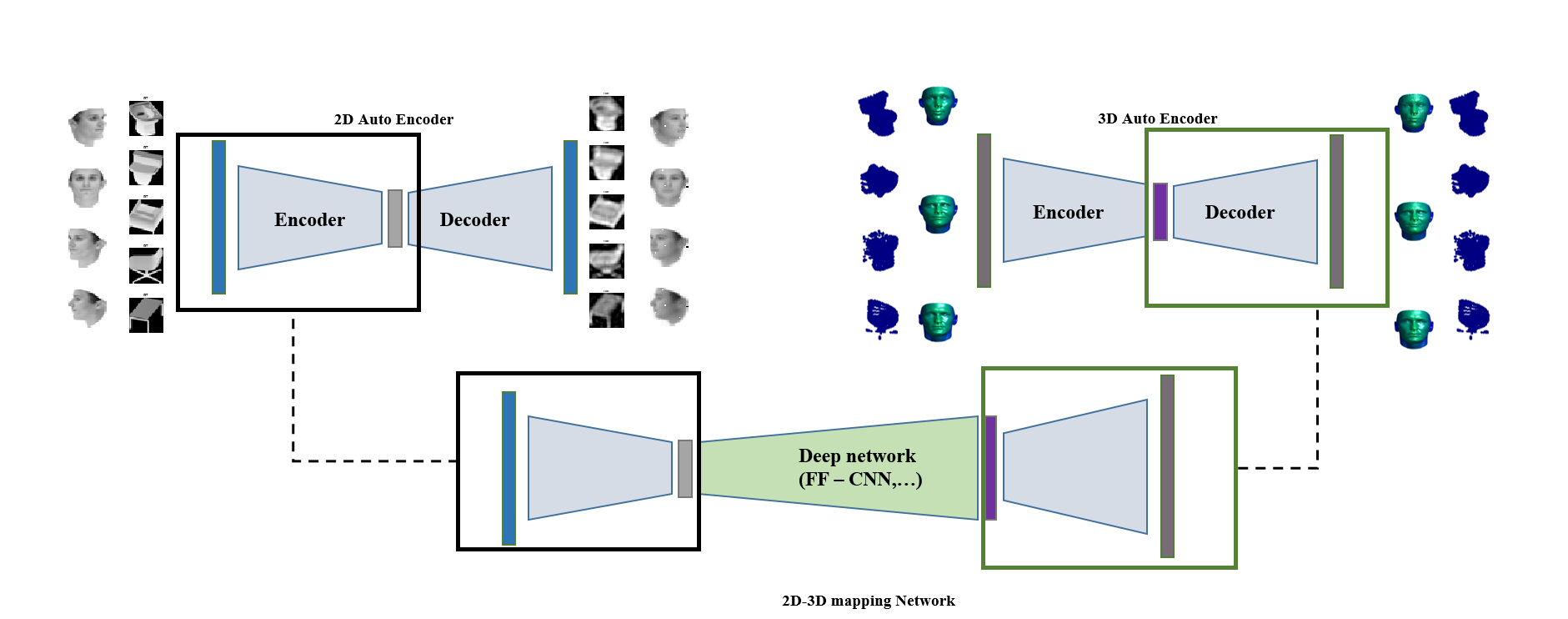}
\caption{\label{fig:f1}Main configuration of proposed framework}
\end{figure}

In the case of using autoencoders for finding representations, inspired by 3D Morphable Models (3DMMs) \cite{blanz1999morphable} the network extracts the basis vectors from training data of each size and uses these basis functions to reconstruct more accurate 3D shape of an object rather that using a pre-defined 3DMM.
Figure \ref{fig:f1} shows the overall structure of the proposed framework. More details about the proposed idea can be found in section \ref{s3}
\section{Related works}\label{s2}
In this section we review and analyze some of the most recent attempts for 3D inverse rendering from single 2D images using deep frameworks as related works to ours concentrating on their search strategies.\par
In \cite{richardson20163d}, Richardson et al. proposed a deep iterative framework for 3D inverse rendering of human faces. In each iteration of this framework, the output of the network will be fed into the input as additional information to fine-tune the previous result. The training process in this method is supervised by using a 3DMM for generating synthetic labeled training data. The final output will be more fine-tuned using some form of Shape From Shading (SFS) method to achieve more details.The authors improve their work in \cite{richardson2016learning} where, the SFS component of their network is implemented as another deep network. The output of this framework is the coefficients of a predefined 3DMM computed using a few hundred 3D samples and so this framework does not directly deal with 3D shape structures.\par
In \cite{tran2016regressing}, a supervised framework for 3D reconstruction of human faces from a single 2D image is proposed where they also used a 3DMM for generating training data. For this aim they fitted a 3DMM to 2D images  via an optimization method and used that model as ground truth for the training data. This strategy of finding ground truth may cause the found solutions to be dependent on the power of the optimization method so that some powers of new work may be ignored. A loss function is also used for the training the network for human faces which can be tuned to make trade-off between generic face generation and overdetermined one, which is suitable for the purpose of face recognition.\par
In \cite{rezende2016unsupervised} an unsupervised framework for 3D reconstruction of objects from single or multiple input images or volumes is proposed. The authors suggested an inference network for encoding the input data to a low dimensional representation which will be fed into a generative network for generating 3D objects using the encoded data. The obtained 3D object will be rendered as a 2D image using a fixed differentiable rendering method. The objective of training of this network is to minimize the difference between rendered and input images. Using a fixed rendering component may restrict the accuracy of generated 3D shapes if the input image rendering mechanism differs from network rendering mechanism.The authors also suggested using a convolutional network as rendering component. Using another deep network as rendering component may overcome the mentioned limitation.\par
In \cite{tewari2017mofa} an autoencoder network is proposed for 3D reconstruction of human faces using an unsupervised training mechanism. In this network the encoder part of the autoencoder is a convolutional network like well-known alexnet \cite{krizhevsky2012imagenet} or  VGG-face \cite{parkhi2015deep} networks and the decoder part is an analytical differentiable formula which uses the encoded data and some assumptions about the image formation process to form to a 3D face and then render it to a 2D image. In this paper the objective is also to minimize the difference between input image and rendered 3D shape using fix rendering method. \par
In \cite{gadelha20163d} a Generative Adversarial Network (GAN) is trained for generating 3D shapes of different objects. In this method, in an unsupervised training approach, a random vector from a distribution will be fed to the network and a 3D shape of an object will be generated and then will be rendered in the form of a 2D image as input to a discriminator. In this way, the discriminator does not need to 3D shape to control the process of learning 3D shapes. The rendering method here is fixed and this network can be used to generate training data for inverse rendering frameworks of 3D objects.\par
In \cite{jackson2017large} a CNN is designed to recover the 3D geometry and texture of human faces from a single 2D image. This framework directly works with 3D coordinates in the form of proposed binary volumes and maps 2D images directly to the coordinates of binary volume. Using binary volume as 3D coordinates for representing human face may result in blur shapes. \par 
 
In this paper, our aim is to design a deep framework for 3D shape inverse rendering of different types of 3D shape representations such as point clouds and volume representations from a single 2D input image. The training process utilizes from both labeled and unlabeled realistic data for improving the quality of resulted 3D shapes. The following section demonstrates the proposed approach in more details.\par

\section{Proposed framework}\label{s3}
The main idea in our work is to end-to-end mapping of some representation of a single 2D input image to some representation of corresponding 3D shape structure. Defining such structure is actually based on the main functionality of deep networks, i. e. using different layers of nonlinear computations in order to feature extraction and transformation \cite{yu2014automatic}. Our proposed framework is also made of components for feature extraction and transformation. \par
As stated in section 2, we used three main components in our framework: the first component is trained to compute some representations for 2D input images. the second component is trained  for reconstructing a 3D shape from some other representations and the third components is used to map the representation computed in the first component to one used in second component. The representations in each component should satisfy its objective. for instance, the component for computing representation from 2D image, should computes a representation that is suitable for being mapped to 3D representation. On the other hand, the 3D representation should be suitable for reconstructing the 3D shape in the form of point cloud or binary volume. We observed in our experiments that more complex representations for 2D representation component achieve better performance and decoder part of a linear autoencoder as 3D reconstruction component is suitable for reconstructing the 3D shape in he form of point cloud. \par
The third component is a structure for mapping 2D low-dim representation to corresponding 3D representation. We actually proposed this component because the manifold of 2D images may be different from the manifold of 3D shapes especially in the case of complicated shapes like human faces and therefore a nonlinear mapping function should be found to obtain an accurate 3D representation from 2D image representation. \par
Note that our proposed frameworks deals directly with dense 3D shapes of the objects instead of using predefined 3D models for representing a face. This characteristic has the advantage of not being restricted to a spanning obtained by a few samples by using larger set of training data for extracting suitable basis vectors for 3D and 2D shape spaces. \par
The main contributions of this paper are as follows:
\begin{itemize}
  \item Using deep structures as different components in an interpretable framework for mapping the representations from a manifold to another and directly reconstructing 3D results. Using deep structures in each component improves handling the non-linearity of each stage.
  \item Possibility of unsupervised pre-training of each component using unlabeled data so that the representations can be improved using realistic data in-the-wild and therefore reducing the need for labeled data for training because of suitable initialization for training. In fact the labeled data is needed only for the final aggregation of pre-trained components.\par
  \item Using the training data itself to extract the bases of each representation space instead of using a predefined analytical models.
  \item Ability of training the framework for inverse rendering of different object classes from a single 2D input image.
\end{itemize}
In the following subsection, the proposed idea in this paper is demonstrated in the simplest case (linear) in terms of Singular Value Decomposition (SVD).
\subsection{Describing proposed idea in a simple case (i.e. linear) in terms of Singular Value Decomposition (SVD)}{\label{4.1}}
In this section we analytically describe the proposed idea in this paper in the simplest case, i.e. the linear case. These calculations gives a mathematical insight into the motivation of proposing this framework. In section \ref{s4}, we will evaluate our idea using different datasets and compare it with other possible method in the case of solving 3D shape inverse rendering problem. 
For this aim, assume the linear case for each component, i. e. linear autoencoder for 2D encoder and 3D decoder components and linear feedforward network as the mapping component. \par
In \cite{baldi1989neural} it is proven that, the optimal representation space resulted by a linear autoencoder, using L2-norm as loss function, has the same spanning as the sub-space resulted by performing PCA on the data. Therefore we can consider the use of a linear autoencoder equivalent to performing PCA on training data. It is therefore can be inferred that the weights of hidden to output layer of the trained linear autoencoder are in fact the bases of the resulted representation subspace which are found using the training data of autoencoder. This characteristic means that, by using a linear autoencoder, we can obtain the bases of a representation space for data of any size without the concern of basis vector decomposition of a large data matrix. \par
Let X denotes the matrix which is formed by concatenating n1 samples of 2D input images vectorized as D-dimensional column vectors as \ref{eq2}:
\begin{equation} \label{eq2}
X=[x_1,...,x_{n_1}]_{D \times n_1}
\end{equation}

By applying SVD on X, after subtracting its mean value, the bases (Eigen vectors) of the PCA subspace can be obtained using SVD as \ref{eq3}:
\begin{equation} \label{eq3}
[U_x,\sigma_x,V_x]=SVD(X)
\end{equation}
where, $U_x$ is the matrix with the columns representing the eigenvectors of $XX^T$, $\sigma_x$  is a diagonal matrix with eigenvalues of $ XX^T $ as diagonal elements and $V_x$ denotes the matrix with the eigenvectors of $X^TX$ on its columns. Therefore we can use the columns of $U_x$  as the basis vectors of the resulted representation subspace for 2D images, and the k-dimensional representation of X in the resulting sub-space, spanned by the first k eigenvectors in $U_x$ with largest eigen values in $\sigma_x$ , can be expressed as Y in \ref{eq4}:
\begin{equation} \label{eq4}
Y_{k \times n_1}=U^{T}_x X_{D\times n_1}
\end{equation}

Similarly, considering Z as the column matrix containing $n_2$ p-dimensional vectorized training samples of 3D shape structures, represented as 3D point clouds or binary volumes, the resulted k’-dimensional representation of Z using another autoencoder can be expressed as $B_{k' \times n_2}$ in \ref{eq7} : 
\begin{equation} \label{eq5}
Z=[z_1,...,z_{n_2}]_{p \times n_2}
\end{equation}
\begin{equation} \label{eq6}
[U_z,\sigma_z,V_z]=SVD(Z)
\end{equation}
\begin{equation} \label{eq7}
B_{k' \times n_2}=U^{T}_z Z_{k'\times p}
\end{equation}

To find a transformation between two obtained representations Y and B for the inverse rendering problem, if a linear feedforward net is used, it is equivalent to performing a least square mechanism to find a linear relation between Y and B as \ref{eq8}:
\begin{equation} \label{eq8}
T_{k \times k'}=argmin_T |B - TY|
\end{equation}
and the result can be used for reconstructing the 3D shape, say $\hat{z}$ ,  as \ref{eq9}:
\begin{equation} \label{eq9}
\hat{z}=U_zTY
\end{equation}
On the other hand, we can also solve a direct optimization problem that linearly tries to find the linear relation between 2D images and corresponding 3D shapes as  \ref{eq10}-\ref{eq11}.
\begin{equation} \label{eq10}
\hat{B}=argmin_B |Z-BX|
\end{equation}
\begin{equation} \label{eq11}
\hat{z}=\hat{B}X
\end{equation}
These two approaches look for a subspace that best describes the 3D shape of the objects in 2D image. The first approach first tries to find a low dimensional representation of 2D input image and corresponding 3D structure for a set of training data and then a linear solution is found for transforming the 2D representation to 3D one via least squares. Note that the matrices $\hat{b}$and $U_{z}T$ are used as mapping functions for the test data. And the examples used for calculating $U_x$ do not necessarily correspond to those which were used to obtain $U_z$ and Y. Such mechanism shows the possibility of unsupervised training of different components in this method.\par
The second closed form approach in \ref{eq11} directly finds a linear mapping between 2D input image and corresponding 3D structure. 
We believe that the first approach significantly improves the process of finding a solution for inverse rendering problem and in Table 2 in the section \ref{5.4}, we verified this claim experimentally compared with second approach in \ref{eq10}. \par
\subsection{Loss function}
As we mentioned before, the purpose of our proposed framework in this paper is to recover the shape of an object. So we deal with a regression problem. Therefore, we used Root Mean Squared Error (RMSE) as in \ref{eq12} as the loss criterion for training the components in the proposed framework. This criterion is the standard deviation of prediction error and is a standard way to regression analysis \cite{dou2017end, kim2017inversefacenet}.\par 
\begin{equation} \label{eq12}
E_{RMSE} = \sqrt{\dfrac{\sum_{i=1}^{N}|\hat{x_o}^i-x_{gt}^i|^2}{N}}
\end{equation}
Where, $\hat{x_o}^i$  and $x_{gt}^i$  denote the prediction and ground truth of the ith training sample and N stands for the sample size. Here, the sample size stands for the size of the mini batch.
There are different loss functions are proposed to be used for inverse rendering problems in different researches \cite{jackson2017large,tran2016regressing} for different applications like reconstruction or recognition. In \cite{jackson2017large} a binary voxel representation is used for representing the faces and sigmoid cross entropy loss function is used for training the network. In our framework we deal with both mesh representation and voxel representations. So we choose the standard RMSE for training the networks.

\section{Experiments}\label{s4}
In this section we perform some experiments to evaluate the performance of our proposed framework in different 3D inverse rendering scenarios. We used 2 types of datasets, objects and human faces having different 3D shape representations, like point clouds and binary volumes, in order to report and compare the reconstruction results. we also analyse some structures as different components of the framework. \par
\subsection{Datasets}{\label{5.1}}
We report the results of the performance of proposed framework on two types of datasets: objects and human faces. In each case we trained a separate structure using that type of data representation. we also used some unlabeled 2D and 3D datasets in order to unsupervised pre-training the components.\par
In the case of human faces, we used LFW dataset \cite{LFWTech} as 2D unlabeled data for unsupervised pre-training and Besel Face Model \cite{paysan20093d} for generating synthetic data as labeled training data along with the Bosphorus dataset \cite{savran2008bosphorus} as realistic labeled data for training and fine-tuning respectively. Note that for aggregating all components using supervised training, we used faces generated by BESEL Face Model (BFM) having natural expression rendered with 3 poses, i.e. -45, 0, 45 degrees about z axis. Therefore the objective of human face shape inverse rendering in our work is to pose invariant identity recovering of faces from a single 2D image. The 3D faces in our framework are represented as 3D point clouds with 54903 vertices in 3D space which are normalized between 0,1 in each dimension. We generated 4000 training and 1000 test samples in natural expression and rendered 2D images of size 32*32 with mentioned poses as 2D input images.\par
In the case of objects, we used some categories of imagenet \cite{deng2009imagenet} dataset as unlabeled data for pre-training and ModelNet10 \cite{wu20153d} and ShapeNet \cite{chang2015shapenet} datasets as labeled training data which are to the best of our knowledge, well-known datasets for 3D object reconstructions. we used 10 categories for training the networks in the form of $30 \times 30 \times 30$ binary voxel grid and rendered each training sample as a $32 \times 32$ gray scale image from 8 viewpoints (between 0 and 180 about z axis). In the case of linear autoencoders we vectorized the voxel grid as a binary column vector of size 27000.\par
We incorporate well-known alexnet in our experiments for evaluating the use of a deep CNN for computing 2D representations for the face dataset. In this case we rendered and used 2D colored images of the faces in each existing dataset with the size $227 \times 227 \times 3$ which is the standard input size to alexnet framework.

\subsection{Parameter setting}{\label{5.2}}
Table \ref{T1} includes the configuration of each structure used for each type of dataset for the type of used structure. Since we applied inverse rendering on different data types, we had to use different structures for the network components in each case.

\begin{table}
\centering
\begin{tabular}{|c|c|c|c|c|c|c|}
 \multicolumn{4}{c}{Object (binary Volume)} & \multicolumn{3}{c}{Human Face (Point cloud)} 
 \\\hline
 & Linear&Nonlinear&Convolutional&Linear&Nonlinear&Convolutional \\\hline
 Encoder& 1024-60 & 1024 - 100- 60 &\makecell{ $3 \times 3 \times 1 -$ \\ $3 \times 3 \times \times 32-$\\ $3 \times 3 \times 16$}& 1024 - 60 & 1024 - 100 - 60& \makecell{$3 \times 3 \times 1 - $ \\ $3 \times 3 \times 32$ \\ $3 \times 3 \times 16$}
 \\\hline
 Mapping& 60-100-400 & 60-100-400& \makecell{$1 \times 3 \times 1 \times 100 - $\\ $400$} & 60-100-98 & 60-100-98 & \makecell{$1 \times 3 \times 100 - $ \\ $98 $}
 \\\hline
 Decoder & 400-27000 & 400-800-27000& \makecell{$3 \times 3 \times 3 \times 32$ \\ $ 3 \times 3 \times 3 \times 16 $ \\$ 30 \times 30 \times 30$} & 98-160470 & 98-200-160470& -
 \\\hline
\end{tabular}
\caption{\label{T1}Network configuration for each components using different structures}
\end{table}

Note that in the case of object datasets, since there are 10 classes of objects, there were need to layers with larger sizes for representation units.\par
The learning rate was set for each configuration so that it gives better performance. In most cases we uses 0.001 for 1000 epochs and then reduce it to 0.00001 for another 1000 epochs. In the case of mesh representation the learning was slower and needed smaller learning rate for longer training time. Batch size was set to a constant, in most cases equal to 40 for training.
We trained our networks from scratch on waterloo university servers: gpu-pr1-01 and gpu-pt1-01, saspc73, and sharcnet-graham machines which are well-known servers having several gpus suitable for training deep networks.\par 

\subsection{Evaluation mechanism}{\label{5.3}}
We used RMSE as the evaluation criterion on test and validation sets in different parts of our experiments. In the case of face datasets we also showed the RMSE in the form of Heat maps, showing point-wise RMSE of reconstructed face on the ground truth face, as a suitable tool for error visualization in different regions of the reconstructed face.\par
We start the experiments by first reporting the evaluation results of closed form linear solutions described in section 4.1 using different data types. in the next step we will use equivalent deep structure by changing the mapping component to nonlinear structure and show that the results can be significantly improved. As a nonlinear component for encoder part of our framework, we fine-tuned alexnet using our labeled human face dataset and report the results. We then show some reconstruction results on realistic human face dataset and compare our results with some of the most recent methods for 3D reconstruction. \par

\subsection{Results of solving linear closed form formula for inverse rendering}
\begin{figure}
\centering
\begin{tabular}{cccc}
\textbf{input image} ($32 \times 32$) & \textbf{Ground Truth} & \makecell{\textbf{LS using low dim } \\ \textbf{ representation}}&\makecell{\textbf{Directly finding Ls }\\ \textbf{between 2D and 3D}}
\\
\includegraphics[width=0.1\textwidth]{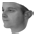}& \includegraphics[width=0.12\textwidth]{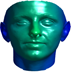}&\includegraphics[width=0.15\textwidth]{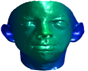}&\includegraphics[width=0.15\textwidth]{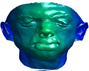}
\\
\includegraphics[width=0.1\textwidth]{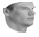}&\includegraphics[width=0.12\textwidth]{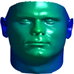}&\includegraphics[width=0.15\textwidth]{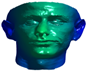} & \includegraphics[width=0.15\textwidth]{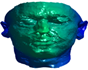}

\end{tabular}
\caption{\label{f2}Visual results of using linear methods for finding a mapping between 2D to 3D spaces for BESE face dataset}
\end{figure}

\begin{figure}[h]
\centering
\begin{tabular}{cccc}
\textbf{input images} ($32 \times 32 each$) & \textbf{Ground Truth} & \makecell{\textbf{LS using low dim } \\ \textbf{ representation}}&\makecell{\textbf{Directly finding Ls  }\\ \textbf{ between 2D and 3D}}
\\
\includegraphics[width=0.2\textwidth]{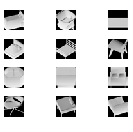}& \includegraphics[width=0.2\textwidth]{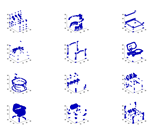}&\includegraphics[width=0.2\textwidth]{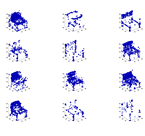}&\includegraphics[width=0.2\textwidth]{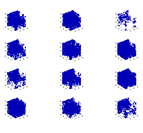}
\end{tabular}
\caption{\label{f3}Visual results of comparing linear least squares for finding a mapping between 2D and 3D space for Modelnet10 dataset}
\end{figure}

{\label{5.4}}
Regarding the analytical description of proposed idea in section \ref{4.1}, here we show that finding a linear mapping between low dimensional representations of 2D and 3D data achieves significantly better results rather that finding a direct linear mapping between them (2D and 3D data). Table \ref{T2} and figures \ref{f2,f3} include the numerical and visual results of applying these methods on different datasets, where “LS using low dim representation” stands for finding the mapping between low dimensional representations of 2D and 3D data using least squares and “Directly finding ls between 2D and 3D” stands for finding a least squares solution for direct mapping 2D input images to 3D data. \par
\begin{table}
\centering
\begin{tabular}{|c|c|c|}
& LS using low dim representation& Directly finding Ls between 2D and 3D
\\\hline
Besel face Dataset & 0.8727 & 2.2869
\\\hline
ModelNet Dataset & 0.0970 & 0.1469 
\\\hline
ShapeNet Dataset & 0.0712 & 0.2455
\\\hline
\end{tabular}
\caption{\label{T2}Comparison of average test RMSE obtained by different datasets using closed form linear solutions in section \ref{4.1}}
\end{table}
The numerical results in Table \ref{T2} indicate the effectiveness of the proposed idea of mapping low dimensional representations instead of direct mapping in terms of RMSE for all existing datasets. 

Looking at Figure \ref{f2}, first, it is clear that using linear methods is not powerful enough for addressing 3D shape inverse rendering problem from a single 2d input image and there is need to use nonlinear tools in this fields especially in the case of human face inverse rendering. And second, we can see that using the low dimensional representation mapping works better that direct mapping visually.\par

In the next step of our experiments, we set up a deep structure consist of linear autoencoders for finding representations and a nonlinear feedforward network for mapping the representations as a strat point for using nonlinearity in the proposed framework.

\subsection{Using linear encoders and decoders with nonlinear feedforward network as mapping component:	}
\begin{figure}[h]
\centering
\begin{tabular}{cccc}
\textbf{input images} ($32 \times 32$) & \textbf{Ground Truth} & Network Reconstruction&Heat Map (RMSE)
\\
\includegraphics[width=0.1\textwidth]{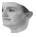}& \includegraphics[width=0.13\textwidth]{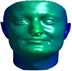}&\includegraphics[width=0.13\textwidth]{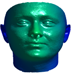}&\includegraphics[width=0.22\textwidth]{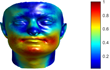}

\\

\includegraphics[width=0.1\textwidth]{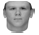}& \includegraphics[width=0.13\textwidth]{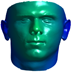}&\includegraphics[width=0.13\textwidth]{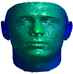}&\includegraphics[width=0.22\textwidth]{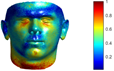}

\\
\includegraphics[width=0.1\textwidth]{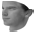}& \includegraphics[width=0.13\textwidth]{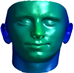}&\includegraphics[width=0.13\textwidth]{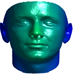}&\includegraphics[width=0.22\textwidth]{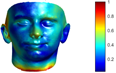}
\\\hline
\multicolumn{4}{c}{Average RMSE (test error)= 0.54611}

\end{tabular}
\caption{\label{f4}Results of using nonlinear feedforward network as mapping component on BESEL face dataset}
\end{figure}

\begin{figure}[h]
\centering
\begin{tabular}{ccc}
\textbf{input images} ($32 \times 32$) & \textbf{Ground Truth} & Network Reconstruction
\\
\includegraphics[width=0.3\textwidth]{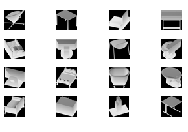}& \includegraphics[width=0.3\textwidth]{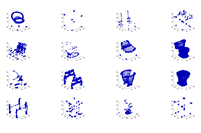}&\includegraphics[width=0.3\textwidth]{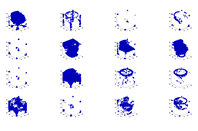}

\\\hline 
\multicolumn{3}{c}{Average RMSE (test error) = 0.0625}

\end{tabular}
\caption{\label{f5}Results of using nonlinear feedforward network as mapping component on Modelnet dataset}
\end{figure}

{\label{5.5}}
As a start point of using nonlinear deep structure in our framework, we only used a nonlinear feedforward network as the mapping component and trained the network using 2000 samples of BESEL face model dataset and training set of Modelnet10 dataset for 1000 epochs. Figures \ref{f4},\ref{f5} show the results on test data for face and object datasets, respectively. \par

It can be observed that using nonlinearity only in the mapping component in our framework results in significant improvement Visually and numerically. For instance, In the case of face datasets, we have face-like reconstructions. this observation also shows that the relation between subspaces of 2D and 3D representations have nonlinear relation. \par

In the next subsection, after fixing the nonlinear feedforward network as mapping component, we consider the use of more complicated representations for 2D and 3D data. To do this, we fine-tuned alexnet as encoder of 2D input image.

\subsection{Transfer learning from Alexnet as 2D encoder}
\begin{figure}[h]
\centering
\includegraphics[width=1\textwidth]{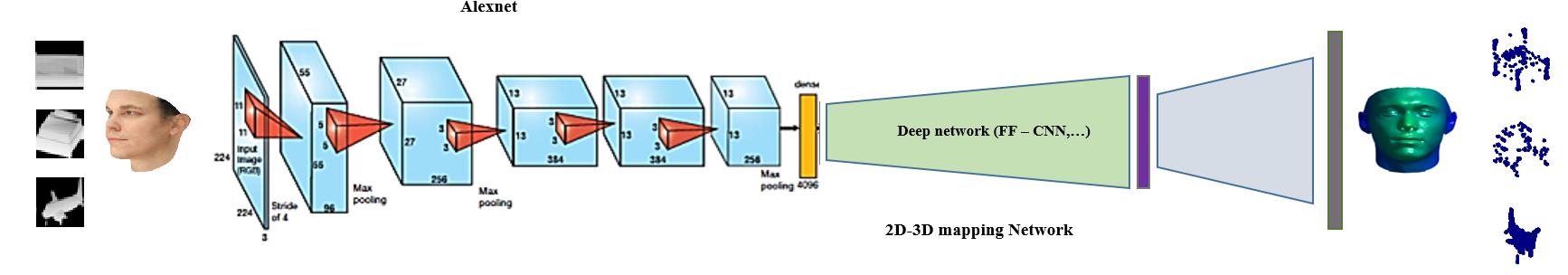}
\caption{\label{f6}alexnet as encoder component in the proposed framework}
\end{figure}

\begin{figure}[h]
\centering
\begin{tabular}{cccc}
\textbf{ \makecell{input image \\($227\times 227\times 3$)}} & \textbf{Ground Truth} & \textbf{Network Reconstruction} & \textbf{Heat Map ( RMSE)}
\\
\includegraphics[width=0.12\textwidth]{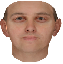}& \includegraphics[width=0.14\textwidth]{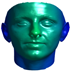}&\includegraphics[width=0.14\textwidth]{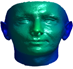}&\includegraphics[width=0.23\textwidth]{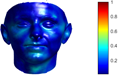}
\\
\includegraphics[width=0.12\textwidth]{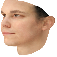}& \includegraphics[width=0.14\textwidth]{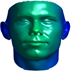}&\includegraphics[width=0.14\textwidth]{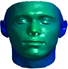}&\includegraphics[width=0.23\textwidth]{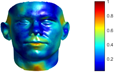}
\\
\includegraphics[width=0.12\textwidth]{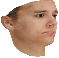}& \includegraphics[width=0.14\textwidth]{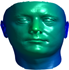}&\includegraphics[width=0.14\textwidth]{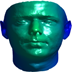}&\includegraphics[width=0.23\textwidth]{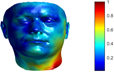}
\\\hline
\multicolumn{4}{c}{Average RMSE = 0.37124}
\end{tabular}
\caption{\label{f7}Visual and numerical results of using alexnet as encoder in proposed framework (BESEL face Dataset)}
\end{figure}

\begin{figure}[h]
\centering
\begin{tabular}{ccc}
\textbf{ \makecell{input image \\($227\times 227\times 3$ each)}} & \textbf{Ground Truth} & \textbf{Network Reconstruction}
\\
\includegraphics[width=0.3\textwidth]{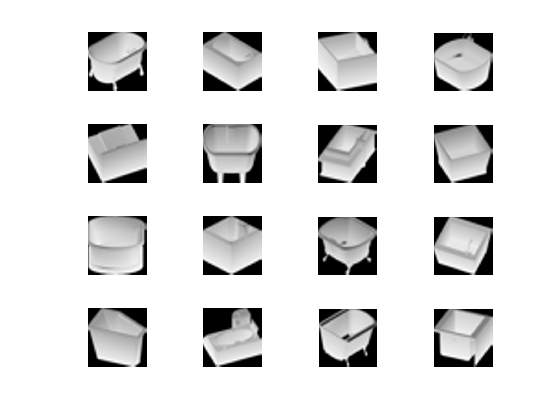}& \includegraphics[width=0.3\textwidth]{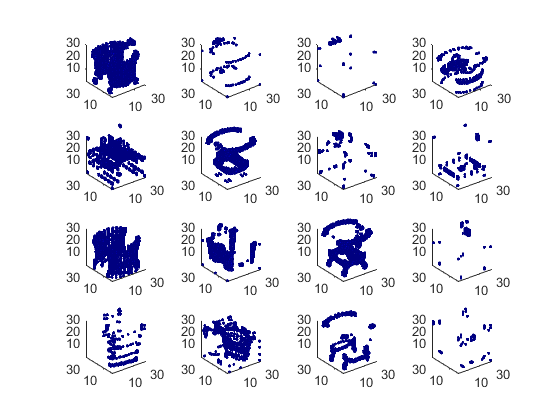}&\includegraphics[width=0.3\textwidth]{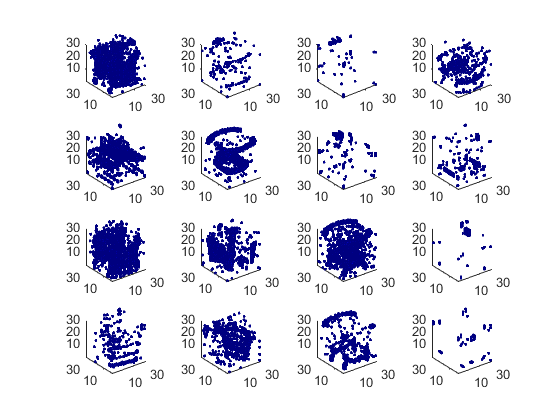}
\\\hline
\multicolumn{3}{c}{Average RMSE = 0.0220}
\end{tabular}
\caption{\label{f7.5}Visual and numerical results of using alexnet as encoder in proposed framework (Modelnet10 Dataset)}
\end{figure}

{\label{5.6}}
Figure \ref{f6} shows the proposed framework which uses Alexnet pre-trained model as encoder component. This model, trained on more than one million images of about 1000 categories of images from ImageNet database \cite{krizhevsky2012imagenet} like human, chair, pencil and many other categories, is a well-known model for extracting rich features from images for classification of them into 1000 object categories.
By using this model as encoder component, we first verify the possibility of using pre-trained components in our framework and second analyze the use of more complicated representations for solving inverse rendering problem in our proposed framework.
We fine-tuned alexnet using a set of colored 2D input images of about 2000 samples generated by BESEL face dataset, and 10000 samples from Modelnet10 and the results are shown in Figures \ref{f7}, \ref{f7.5} respectively. in the case of Modelnet10, Since there were no texture associated with the object to be rendered as a color image, we just represent the input images with the size $227 \times 227 \times 3$ to feed to the network.

From Figures \ref{f7}, \ref{f7.5} it is observable that using this configuration and with computing more complex representations for 2D input images, the accuracy of reconstruction will be improved compared with the linear autoencoder. Actually, the objective of encoder component is to compute representations which are more suitable for being mapped to 3D representation and the results show that more complicated representations are appropriate for this aim.\par
\begin{figure}[h]
\centering
\begin{tabular}{cc}
\includegraphics[width=0.5\textwidth]{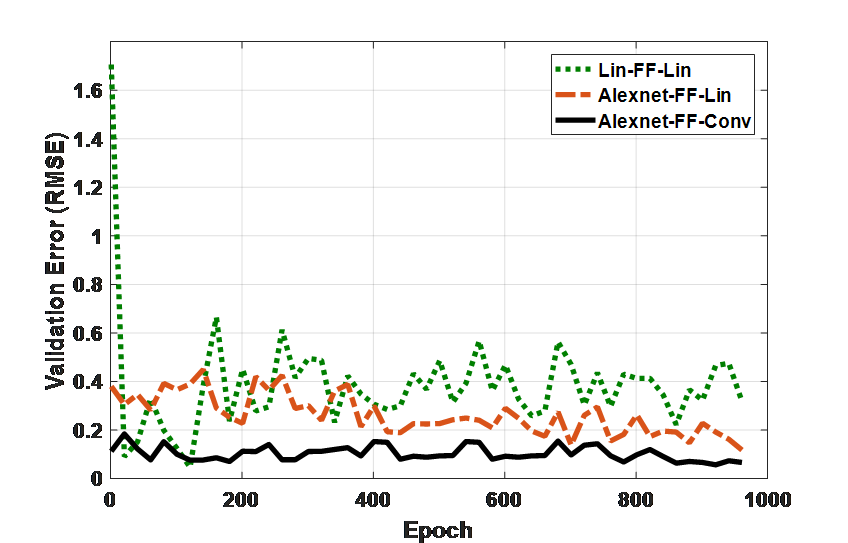}& \includegraphics[width=0.5\textwidth]{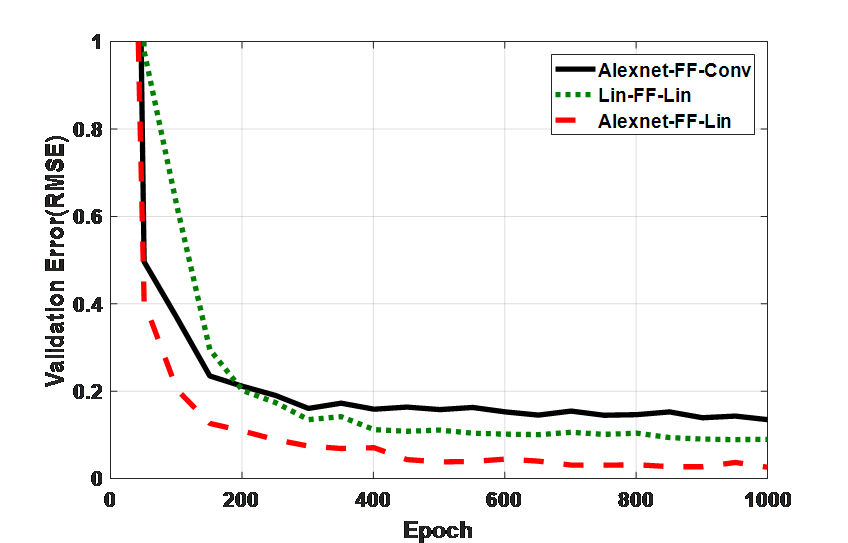}
\\ (a) ModelNet Dataset & (b) BESEL Face Dataset
\end{tabular}
\caption{\label{f8} Validation error vs Epoch for Different structures used as components in proposed frameworks for (a) ModelNet dataset and (b) Face dataset where Lin: Linear encoder or decoder, FF: Nonlinear FeedForward network, Conv: Convolutional encoder or decoder}
\end{figure}

In the case of decoder component, the aim is recovering sharp 3D shapes from 3D representations. therefore, the representation in this component should be so that the 3D shape can be inferred from. in the case of point cloud 3d shapes, where we directly deal with real 3D coordinate values the linear decoder gives the most promising result compared with nonlinear and convolutional structures with blur results. in the case of binary volume 3D shapes, where the output can be digitized to 0 and 1 using some threshold value, nonlinear structures show promising performances. figure \ref{f8} shows the digaram of RMSE vs epoch using different mentioned structures for different types of 3D shapes.\par

\subsection{Pose invariant reconstruction }
\begin{figure}[h]
\centering
\begin{tabular}{ccc}
Input image& Network Reconstruction & Heat map (RMSE)
\\
\includegraphics[width=0.12\textwidth]{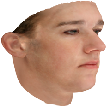}& \includegraphics[width=0.12\textwidth]{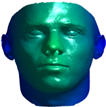}&\includegraphics[width=0.2\textwidth]{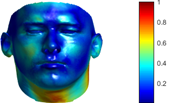}
\\
\includegraphics[width=0.12\textwidth]{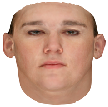}& \includegraphics[width=0.12\textwidth]{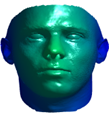}&\includegraphics[width=0.2\textwidth]{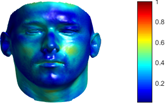}
\\
\includegraphics[width=0.12\textwidth]{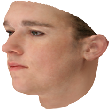}& \includegraphics[width=0.12\textwidth]{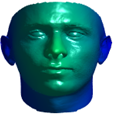}&\includegraphics[width=0.2\textwidth]{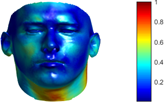}
\end{tabular}
\caption{\label{f8.5}Result of proposed framework on Bosphorus face dataset}
\end{figure}

{\label{5.5.1}}
Figure \ref{f8.5} shows the power of proposed framework in reconstruction of a sample face input image captured from 3 different poses. 
\subsection{Reconstruction results on realistic datasets}{\label{5.5}}
We evaluated the final configuration of our framework (Alexnet encoder – nonlinear mapping – linear decoder) on Bosphorus Face dataset as a realistic 3D face dataset. Figure 8 shows the results and their heat map in terms of RMSE on Bosphorus dataset.We also compared the visual reconstruction results of proposed framework with \cite{jackson2017large} and \cite{tewari2017mofa} on some realistic images.\par
 In the case of realistic images we first manually cropped the face from input image and then aligned tha intensity of the image with a sample image from training samples of our framework using Procrustes analysis used in \cite{skoglund2003three} and then use them as input image to our framework. Note that in order to compute RMSE for heat map we computed the minimum RMSE from Ground truth to network reconstruction.\par
\begin{figure}[h]
\centering
\begin{tabular}{cccc}
\includegraphics[width=0.10\textwidth]{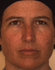}& \includegraphics[width=0.13\textwidth]{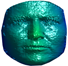}&\includegraphics[width=0.13\textwidth]{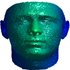}&\includegraphics[width=0.21\textwidth]{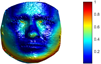}
\\
\includegraphics[width=0.1\textwidth]{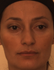}& \includegraphics[width=0.13\textwidth]{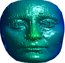}&\includegraphics[width=0.13\textwidth]{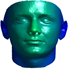}&\includegraphics[width=0.21\textwidth]{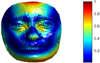}
\\
\includegraphics[width=0.1\textwidth]{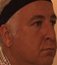}& \includegraphics[width=0.13\textwidth]{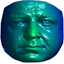}&\includegraphics[width=0.13\textwidth]{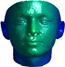}&\includegraphics[width=0.21\textwidth]{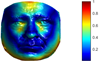}
\\
\includegraphics[width=0.1\textwidth]{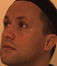}& \includegraphics[width=0.13\textwidth]{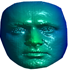}&\includegraphics[width=0.13\textwidth]{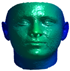}&\includegraphics[width=0.21\textwidth]{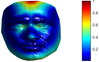}
\end{tabular}
\caption{\label{f9}Result of proposed framework on Bosphorus face dataset}
\end{figure}
The results in Figure \ref{f9} show that the network could successfully recover the main features in each face. Note that using the linear decoder in the case of face datasets is equivalent to performing PCA on all of training samples (about 2000) and this causes achieving better results compared with a predefined 3DMM using small set of training data.
Figure 9 shows some visual reconstruction results, using some realistic images, obtained by proposed framework compared with two similar recent reconstruction methods \cite{tewari2017mofa} and \cite{jackson2017large} which where described in section \ref{s2}. We can see that the results obtained by proposed framework are comparable with unsupervised raining methods and can detect more shape details compared with the binary volume representation for 3D shapes. We believe this is because of dealing directly with point clouds and extracting basis vectors from a large set of training data.

\begin{figure}[h]
\centering
\begin{tabular}{cccc}
Input image & Tewari et al. \cite{tewari2017mofa} & Jackson \cite{jackson2017large} & Ours
\\
\includegraphics[width=0.12\textwidth]{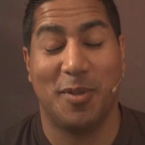}& \includegraphics[width=0.12\textwidth]{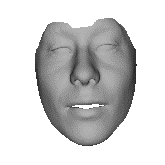}&\includegraphics[width=0.1\textwidth]{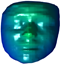}&\includegraphics[width=0.11\textwidth]{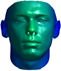}
\\
\includegraphics[width=0.12\textwidth]{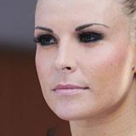}& \includegraphics[width=0.12\textwidth]{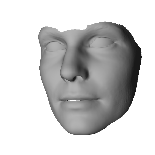}&\includegraphics[width=0.1\textwidth]{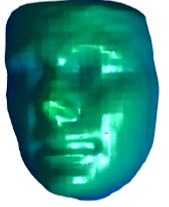}&\includegraphics[width=0.11\textwidth]{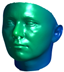}
\\
\includegraphics[width=0.12\textwidth]{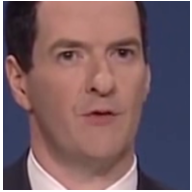}& \includegraphics[width=0.096\textwidth]{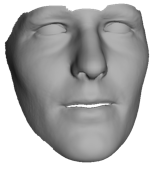}&\includegraphics[width=0.1\textwidth]{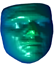}&\includegraphics[width=0.11\textwidth]{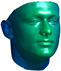}
\\
\includegraphics[width=0.12\textwidth]{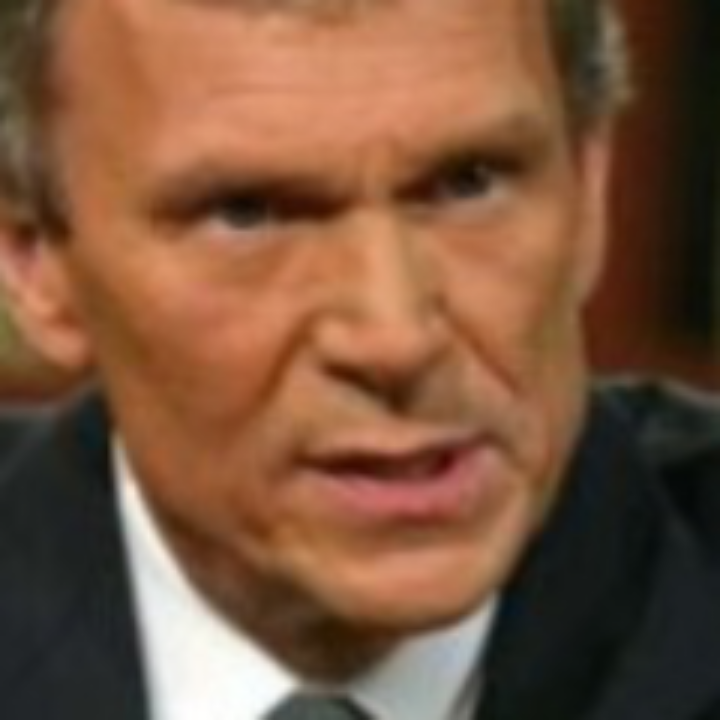}& \includegraphics[width=0.12\textwidth]{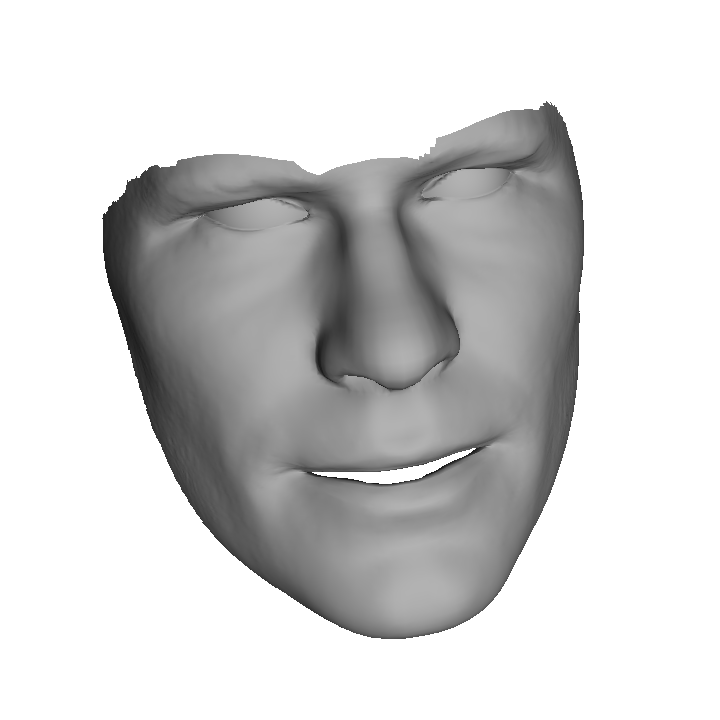}&\includegraphics[width=0.1\textwidth]{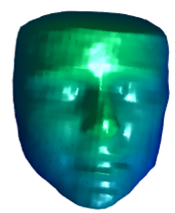}&\includegraphics[width=0.11\textwidth]{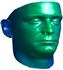}
\end{tabular}
\caption{\label{f10}Comparing viual reconstruction results with \cite{jackson2017large,tewari2017mofa}}
\end{figure}
\section{Discussion}\label{s5}
As we observed in the experiments, using a convolutional network as 2D encoder and mapping component in our framework significantly improves its performance. However in the case of 3D decoder using a non-convolutional network (linear or non-linear) results in better reconstruction results. We believe this is because a in convolutional Autoencoder, reconstruction of the representation obtained from data may be inaccurate because of the information loss resulted by convolution and pooling units. \par
Another Property of our proposed framework is the use of autoencoder to directly reconstruct the 3D shape structure by learning from large set of examples. this structure Uses the training data itself to extract bases of the shape space. We believe and this characteristic leads to have unbiased reconstructions rather than using a predefined 3D model which is biased toward mean shape of the training samples. On the other hand, working directly with point cloud representation for representing the 3D shape helps the framework to achieve more detailed results.\par
As a disadvantage of regressing process in our framework, we can mention the existing loss functions. in the case of situation which the output of the network should be some exact values for regression, there is need to define a suitable loss function to guide the learning process toward finding desired solution. If we can turn the problem of regression into the form of some classification problem using losses like cross entropy and so on, we will achieve more desirable solutions like the results obtained by Generator Adversarial Netwok (GAN) frameworks.\par
\section{Conclusions and future works}\label{s6}
In this paper, a semi-supervised interpretable framework is proposed for pose invariant 3D shape inverse rendering from a single 2D input image. Using autoencoder based on mapping suitable low dimensional representations computed from 2D and 3D data components for extracting representations and reconstruction of final 3D shapes in the proposed framework, on one hand, makes it possible to use unlabeled data for unsupervised pre-training and therefore reducing the need to large sets of labeled data for training the framework and on the other hand, helps to achieve more promising and detailed reconstructions because of extraction of bases vector of 3D shape space from the large set of training data compared with predefined 3D models.\par
in this proposed framework two types of 3D shape representations were used for reconstruction: point cloud and binary volumes. The obtained results show that binary volumes are more suitable representations for more complicated deep networks. However they express blur shapes with high dimensional representation and can express less details compared with point cloud representations. on the other hand, working with exact 3D coordinates of point clouds is not too easy for deep networks and linear structures are preferred for such shape reconstructions. \par
In the future attempts one can unify the Inverse rendering framework for working with both point cloud and volume shape structures for different class of objects.\par

\bibliographystyle{acm}
\bibliography{sample}

\end{document}